\documentclass[10pt,twocolumn,letterpaper]{article}

\usepackage{wacv}
\usepackage{times}
\usepackage{epsfig}
\usepackage{graphicx}
\usepackage{amsmath}
\usepackage{amssymb}
\usepackage{booktabs}
\usepackage{commath}
\usepackage{multirow}
\usepackage[table,xcdraw]{xcolor}
\usepackage{pifont}
\usepackage[accsupp]{axessibility}
\def\wacvPaperID{887} 
\wacvfinalcopy 

\ifwacvfinal
\usepackage[breaklinks=true,bookmarks=false]{hyperref}
\else
\usepackage[pagebackref=true,breaklinks=true,colorlinks,bookmarks=false]{hyperref}
\fi

\begin{document}

\title{DE-CROP: Data-efficient Certified Robustness for Pretrained Classifiers}

\author{
\vspace{-0.15in}
\begin{tabular}{c c c}
\centering
  \hspace{0.4in}Gaurav Kumar Nayak\thanks{denotes equal contribution.} & \hspace{-0.9in}Ruchit Rawal\footnotemark[1] & \hspace{-0.9in}Anirban Chakraborty  \vspace{0.1in}\tabularnewline
  & \hspace{-0.9in}Department of Computational and Data Sciences \tabularnewline
& \hspace{-0.9in}Indian Institute of Science, Bangalore, India \tabularnewline &
\hspace{-0.9in}{\tt\small \{gauravnayak, ruchitrawal, anirban\}@iisc.ac.in}
\end{tabular}\\
}
\maketitle
\thispagestyle{empty}

\begin{abstract}
   Certified defense using randomized smoothing is a popular technique to provide robustness guarantees for deep neural networks against $l_{2}$ adversarial attacks. Existing works use this technique to provably secure a pretrained non-robust model by training a custom denoiser network 
   on entire training data. 
   However, access to the training set may be restricted to a handful of data samples 
   due to constraints such as high transmission cost and the proprietary nature of the data. Thus, we formulate a novel problem of “how to certify the robustness of pretrained models using only a few training samples”. We observe that training the custom denoiser directly using the existing techniques on limited samples yields poor certification. To overcome this, our proposed approach (DE-CROP)\footnote{Project Page: \href{https://sites.google.com/view/decrop}{https://sites.google.com/view/decrop}} generates class-boundary and interpolated samples corresponding to each training sample, ensuring high diversity in the feature space of the pretrained classifier. We train the denoiser by maximizing the similarity between the denoised output of the generated sample and the original training sample in the classifier's logit space. 
   We also perform distribution level matching using domain discriminator and maximum mean discrepancy that yields further benefit. In white box setup, we obtain significant improvements over the baseline on multiple benchmark datasets and also report similar performance under the challenging black box setup.
\end{abstract}


\vspace{-0.15in}
\section{Introduction}
\label{sec:intro}
Given large amounts of training data (such as ImageNet~\cite{deng2009imagenet}) and compute power (powerful GPUs~\cite{dl_gpu}), deep models are highly accurate on the respective tasks for which it is trained. However, these models can easily get fooled when they encounter adversarial images as input that are crafted using an adversarial attack~\cite{szegedy2013intriguing}. 
A lot of efforts have been made in the literature to secure the models against adversarial attacks. Adversarial training~\cite{madry2017towards,goodfellow2014explaining} is one of the most common ways to provide empirical defense where the models are trained to minimize the maximum training loss induced by adversarial samples. Such defenses are heuristic-based and are only robust to known or specific adversarial attacks. Powerful adversaries easily break them, hence are not truly robust against adversarial perturbations~\cite{athalye2018robustness,uesato2018adversarial,croce2020reliable}. This motivated the researchers to develop methods where a trained model can be guaranteed to have a constant prediction in the input neighborhood. Such methods that provide formal guarantees are called certified defense methods~\cite{raghunathan2018certified,wong2018provable,ehlers2017formal,katz2017reluplex,mirman2018differentiable}. 

Randomized smoothing~\cite{cao2017mitigating, liu2018towards,lecuyer2019certified,li2019certified,cohen2019certified} is a certified defense technique used to provide provable robustness against $l_{2}$ adversarial perturbations. It outperforms other certification methods and is also scalable to deep neural networks due to its architectural independence. Using randomized smoothing, any base classifier can be converted to a smoothed classifier, which is certifiably robust against $l_{2}$ attacks as it has the desirable property of constant predictions within the $l_{2}$ ball around the input. The prediction by the smoothed classifier on an input image is simply the most probable class predicted by the base classifier on random gaussian perturbations of the input. Note that the higher the probability with which the base classifier predicts the most probable class as the correct class, the higher the certified radius~\cite{cohen2019certified}. However, any vanilla trained / off-the-shelf classifier would not be robust to the input corrupted by gaussian noise. The predicted most probable class can be incorrect or may be predicted with very low confidence leading to poor certification. Thus, models are often trained from scratch using the gaussian perturbed samples~\cite{cohen2019certified} as augmentation, and also with adversarial training~\cite{salman2019provably}.

\begin{figure}[htp]
\centering
\centerline{\includegraphics[width=0.49\textwidth]{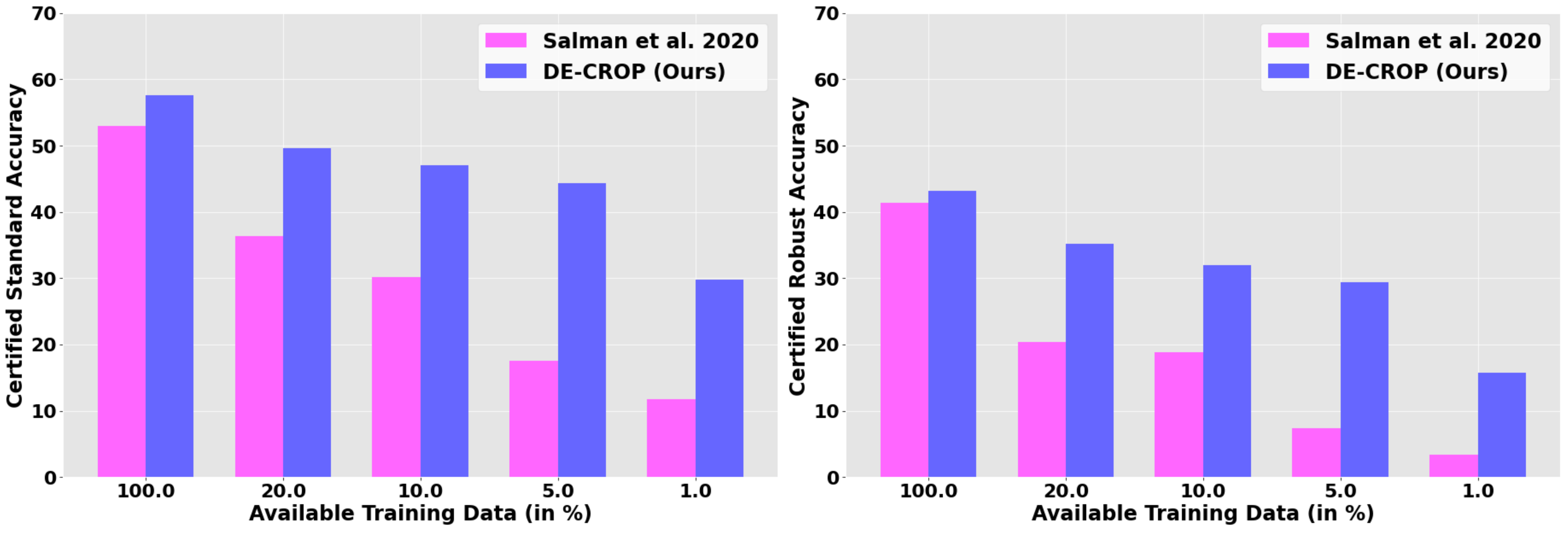}}
\caption{The certified accuracy (both standard and robust) of existing method drops significantly as the training set size of CIFAR-$10$ reduces. Our method (\textit{DE-CROP}) overcomes this issue 
by obtaining significant gains in certified performance against $l_2$ perturbations within radius $0.25$, across different training data budgets. We use gaussian noise with $\sigma$ as $0.5$ for certification.
}
\label{fig:motivation}
\end{figure}

Training from scratch on gaussian-noise augmentation is not always a viable option, especially when the large pretrained models are shared as an API, either as a white box or a black box. Also, retraining these large cumbersome models adds a lot of computational overload. To avoid this, Salman \etal~\cite{salman2020denoised} proposed a “denoised smoothing” technique where a custom-trained denoiser is prepended before the pretrained classifier. Though their approach provides certified robustness to pretrained models, they use the entire training data for training the denoiser. In fact, all the previous certification methods also assume the availability of entire training data. This assumption is unrealistic as the API provider may not share the whole large-scale training set. Due to heavy transmission costs associated with complete training data or proprietary reasons, they may provide access to only a few training samples. In such cases, when we perform denoised smoothing
~\cite{salman2020denoised} directly, the certified accuracy decreases significantly as the number of training samples reduces from $100\%$ to $1\%$ as shown in Fig.~\ref{fig:motivation}. 

In this work, we attempt to provide certified robustness to a pretrained non-robust model in the challenging setting of limited training data. We first explore whether the performance of denoised smoothing can be improved using different approaches such as weight decay as a regularization scheme or even generate extra data using augmentation and mix-up strategies \cite{zhang2017mixup} to avoid overfitting. However, these traditional approaches only show a marginal improvement (refer to Tables~\ref{tab:table1},~\ref{tab:table2}). Hence, we propose our approach of \textbf{d}ata-\textbf{e}fficient \textbf{c}ertified \textbf{ro}bustness (dubbed as `\textit{DE-CROP}’)  (refer to Sec.~\ref{sec:proposed} and Fig.~\ref{fig:tsne},~\ref{fig:approach}) that generates better samples, yielding diverse features on the pretrained classifier similar to complete training data.

Since generative models are hard to train and perform poorly (on downstream tasks) in the presence of limited data, we adopt a simple and intuitive approach to generate additional data to minimize overfitting. We achieve this by generating adversarial samples (corresponding to our limited data) that we term `boundary' samples. Adversarial samples act as an upper bound to the decision boundary~\cite{nanda2021fairness} while maintaining the semantic content of the image, allowing the denoiser to learn from samples in its neighborhood. Moreover, we also generate interpolated samples (lying between original and boundary samples) that further increase the data diversity in the feature space. We train the denoiser by maximizing the similarity between the pre-trained classifier’s features on the original data and denoised perturbed generated/original data. We achieve this by performing orthogonal modifications at two granularities: a) instance-level (using cosine-similarity) and b) distribution-level (using Maximum Mean Desprecancy \cite{gretton2012kernel} and negative gradients from a Domain-discriminator \cite{ganin2015unsupervised}). In our experiments, we observed that although cosine-similarity improved denoised performance by exploiting the discrimnativeness of the pretrained classifier, its benefits were limited as it only operates at an instance level. Thus, inspired by fundamental ideas (such as maximum mean discrepancy and domain discriminator) in domain adaptation, we formulate the objective of obtaining correct predictions on denoised images by reducing the distribution discrepancy between feature representations of the original clean inputs and the denoised gaussian perturbed output. As shown in Fig.~\ref{fig:motivation}, our 
method DE-CROP significantly improves certified performance across different sample budgets ($100\%$, $20\%$, $10\%$, $5\%$, $1\%$) on CIFAR-$10$ compared to ~\cite{salman2020denoised}.

Our contributions are summarized as follows:
\begin{itemize}
\item 
Given only 
limited training data, we provide robustness guarantees for a non-robust pretrained classifier against $l_{2}$ perturbations on both white-box and black-box setups. To the best of our knowledge, we are the first to provide certified adversarial defense using only the few training samples.
\item 
To mitigate overfitting on limited training data, 
we propose a novel sample-generation strategy that synthesize 
`boundary' and `interpolated' samples (Sec.~\ref{subsec:generation}) to augment the limited training data, 
leading to improved 
feature-diversity on the pretrained classifier. 
\item The denoiser network trained with regular cross entropy loss provides limited benefit. To enhance the performance further, we proposed additional losses (Sec.~\ref{subsec:denoiser_training}) that align the feature representations of original and denoised gaussian perturbed generated/original samples at multiple granularities (both instance and distribution levels).
\item We show benefit of our 
generated samples (Sec.~\ref{gen_data}) along with contributions from each of the proposed losses (Sec.~\ref{dist_align}), by reporting significant improvements observed across 
diverse sample budgets and noise levels in both white-box 
and black-box settings. 
\end{itemize}
\vspace{-0.1in}
\section{Related Works}
\label{sec:related}
We broadly categorize the relevant works that provides adversarial robustness 
and briefly discuss them below:

\textbf{Empirical Robustness}: Empirically motivated adversarial robustness defenses can be broadly classified into: a) adversarial training (AT) and b) non-adversarial training regularizations. AT  \cite{madry2017towards, goodfellow2014explaining, szegedy2013intriguing} improves robustness by augmenting the training data with adversarial samples generated by a particular threat model. Although AT is widely regarded as the best empirical defense, it suffers from high-computational costs due to generation of adversarial samples at training time. Non-AT based approaches \cite{chan2019jacobian} attempt to reduce the computational burden (usually at the cost of drop in adversarial robustness) by mimicking properties observed in robust networks explicitly. 
AT is also highly dependent on the quantity of training data. Aditi \etal \cite{carmon2019unlabeled} demonstrated that using additional unlabeled data in a pseudo-labelling setup led to a significant increase in adversarial robustness. 
However, since performance of pseudo-labelling itself  depends on the amount of labeled data, the performance of their technique drops considerably as labeled data is reduced. To alleviate this problem, Sehwag \etal \cite{sehwag2021robust} illustrated the benefit of using additional data generated from generative models for improving adversarial robustness. 

Since the empirical defenses are designed based on heuristics, they can be easily fooled as stronger adversarial attacks are developed in the future. In contrast, we attempt to provably robustify pretrained classifier against $l_{2}$ adversarial attacks under the challenging constraint of limited training samples.

\textbf{Certified Robustness}: Unlike empirical defense, here the model predictions on the neighborhood region lying within ball of small radius around the input sample, is guaranteed to not vary and remain constant. The methods that provide certification are either ‘exact’~\cite{fischetti2017deep,ehlers2017formal,katz2017reluplex,tjeng2019evaluating} or ‘conservative’~\cite{wong2018provable,zhang2018efficient}. The former  is not scalable to large architectures, is compute-intensive, and often uses less expressive networks but can verify with guarantees about the existence of any adversarial samples if lying within the radius of input. The latter is more scalable and requires less computation but can incorrectly decline the certification even if no adversarial sample is present. Both techniques require either customized or specific architectures, hence are not suitable for modern deep architectures. 

Randomized smoothing is a popular technique that does not have any architectural dependency and was initially used as heuristic defense~\cite{cao2017mitigating, liu2018towards}. It was first shown to provide certified guarantees by Lecuyer \etal~\cite{lecuyer2019certified} where techniques from `differential privacy’ were used for certification. It was later improved by Li \etal~\cite{li2019certified} who provided better guarantees using ideas from `information theory’. Both these methods have lower guarantees on the smoothed classifier. Cohen \etal~\cite{cohen2019certified} provided a tight certified guarantee against $l_{2}$ norm adversarial perturbations. After that, the certified accuracy was further improved by using adversarial training techniques in the randomized smoothing framework~\cite{salman2019provably}. However, all these techniques trained the classifier from scratch while providing certified robustness. Recently, Salman \etal~\cite{salman2020denoised} provided provable robustness to pretrained models by appending a custom trained denoiser before it. We also train the custom denoiser but only using few training samples. Unlike \cite{salman2020denoised}, our limited data setup is more challenging and directly using their method yields poor certification results. Our generated samples with the added domain discriminator optimized using our proposed losses, 
handles the overfitting on the denoiser quite well and gives significant improvements on certified accuracy.

We now discuss the necessary preliminaries to give required background before explaining  our approach. 
\vspace{-0.02in}
\section{Preliminaries}
\label{sec:prelims}
\textbf{Notations}: The complete original training dataset is denoted by $D_o = \{D_{train}, D_{test}\}$, where $D_{train}$ and $D_{test}$ are the training set and the test set respectively. The base classifier $B_c$ is pretrained on the entire $D_{train}$ which consists of $N$ training samples. The API provider has granted public access to the trained $B_c$ which can be used by clients to obtain predictions. However, only a limited amount of $D_{train}$ (denoted by $D_{train}^{lim}$) is shared to the clients. $D_{train}^{lim}$ is only $k\%$ of $D_{train}$ containing $N^k$ training samples such that $N^k \ll N$). The same relationship holds for each class of the classifier $B_c$ i.e. for any class $c$: $N_c^k \ll N_c$ and $N_c^k$ is $k\%$ of $N_c$. An $i^{th}$ sample of $D_{train}^{lim}$ (i.e. $x_o^i$)  with label $y_o^i$ is perturbed by gaussian noise which is denoted by $\bar{x}_o^i = x_o^i + \epsilon$ where $\epsilon \sim \mathcal{N}(0, \sigma^2 I)$, constituting the collection of perturbed samples as $\bar{D}_{train}^{lim}$. 

The penultimate features, logits and label predictions from the pretrained classifier $B_c$ on $x_o^i$ are denoted by $F_{B_c}(x_o^i)$, $label(B_c(x_o^i))$ and $B_c(x_o^i)$ respectively. Similarly, $label(S_c(x_o^i))$ is the label predicted by smoothed classifier $S_c$ on input $x_o^i$. The classifier $B_c$ is converted to smoothed classifier $S_c$, which is used for certified defense via randomized smoothing. The certified radius of an $i^{th}$ test sample is $R_c^i$. The evaluations are performed on $l_{2}$ perturbations at a radius $r$ denoted by $l_2^r$. The denoiser network $D_n$ and domain discriminator $D_d$, are parameterized by $\theta$ and $\phi$ respectively. The boundary sample and interpolated sample generated using an $i^{th}$ training sample of $D_{train}^{lim}$ (i.e. $x_o^i$) are represented by $x_b^i$ and $x_{int}^i$ respectively.

\textbf{Randomized Smoothing (RS)}: This technique is used to build a new smoothed classifier $S_c$ from the given base classifier $B_c$. For any $i^{th}$ sample of $D_{train}^{lim}$ (i.e. $x_o^i$), the output of the classifier $S_c$ corresponding to input $x_o^i$ is the most likely class that has the highest probability to be get predicted by $B_c$ on $\bar{x}_o^i$.
\begin{equation}
\begin{gathered}
\label{eq1}
label(S_c(x_o^i)) = \underset{c \in C}{\mathrm{argmax}}\hspace{0.1in} Prob(label(B_c(x_o^i + \epsilon)) = c)\\ \text{where} \hspace{0.1in} \epsilon \sim \mathcal{N}(0, \sigma^2 I)
\end{gathered}
\end{equation}
Here, $\sigma$ is a hyperparameter which controls the noise level and $C$ is the set of unique target labels in $D_{train}^{lim}$. The process of RS does not assume $B_c$'s architecture and thus permits the $B_c$ to be any arbitrary large deep neural network.

\pagebreak
\textbf{Certified Robustness using RS}: Lecuyer \etal~\cite{lecuyer2019certified} and Li \etal~\cite{li2019certified} gave robustness guarantees for the smoothed classifier $S_c$ using RS, but they were loose as $S_c$ was provably more robust than the obtained guarantees. Cohen~\etal~\cite{cohen2019certified} gave a tight bound on $l_{2}$ robust guarantees using RS.  

If the prediction on the base classifier $B_c$ for the gaussian perturbed copies of $x_o^i$ i.e. $\mathcal{N}(x_o^i, \sigma^2 I)$ is $c_1$ as the ‘most probable class’ with probability $p_1$ and $c_2$ as “runner-up” class with $p_2$, then the smoothed classifier $S_c$ is provably robust around the input $x_o^i$ within the radius $R_c = \sigma/2 (\phi^{-1}(p_1)-\phi^{-1}(p_2))$. Here $R_c$ is the certified radius as the predictions are guaranteed to remain constant inside the radius and $\phi^{-1}$ denotes the inverse of standard gaussian CDF. It is impossible to calculate $p_1$ and $p_2$ exactly when $B_c$ is a deep neural network. Hence, Cohen \etal, estimates lower bound on $p_1$ ($\underline{p_1}$) and upper bound on $p_2$ ($\overline{p_2}$) using Monte Carlo technique.  

\textbf{Theorem} [Cohen \etal~\cite{cohen2019certified}]: \textit{Let $B_c$ be any function that maps the input to one of the output class labels. Let $S_c$ be defined as in eq.~\ref{eq1}. The noise $\epsilon$ is sampled from the normal distribution i.e. $\epsilon \sim \mathcal{N}(0, \sigma^2 I)$. If $c_1 \in C$ and $\underline{p_1}$, $\overline{p_2} \in [0,1]$ holds the below inequality}:
\begin{equation}
\begin{gathered}
Prob(label(B_c(x_o^i+\epsilon))=c_1) \geq \underline{p_1} \geq \overline{p_2} >= \\\underset{c \neq c_1}{max} Prob(label(B_c(x_o^i+\epsilon)=c)
\end{gathered}
\end{equation}
\textit{Then $label(S_c(x_o^i+\epsilon)=c_1 \forall \left\lVert\epsilon\right\rVert_{2} < R_c$, where}
\begin{equation}
\begin{gathered}
\label{eq3}
R_c =   \sigma/2 ( \phi^{-1}(\underline{p_1}) - \phi^{-1}(\overline{p_2}) )
\end{gathered}
\end{equation}
In practice, Cohen \etal used $R_c = \sigma \phi^{-1}(\underline{p_1})$ for $\underline{p_1}>1/2$, assuming $\overline{p_2} = 1-\underline{p_1}$, otherwise abstained with $R_c =0$. These above expressions can be derived using the “Neyman-Pearson” lemma for which we refer the reader to~\cite{cohen2019certified}. Now, we discuss our proposed approach in detail in the next section.

\section{Proposed Approach}
\label{sec:proposed}
We aim to provide certified robustness to the given pretrained base classifier $B_c$. However, obtaining certification using randomized smoothing expects the model $B_c$ to be robust against the input perturbations with the random Gaussian noise, which may not be the case with the model $B_c$ supplied by the API provider. In order to make the base model $B_c$ appropriate for randomized smoothing based certification without modifying/retraining $B_c$, a denoiser network $D_n$ is prepended to $B_c$. Thus, $B_c \circ D_n$ is the new base classifier using which the prediction on smoothed classifier $S_c$ on an $i^{th}$ training sample is defined as follows:
\begin{equation}
\begin{gathered}
\label{eq4}
label(S_c(x_o^i))=\underset{c \in C}{\mathrm{argmax}}\hspace{0.01in}Prob(label(B_c(D_n(x_o^i + \epsilon)))=c)\\ \text{where} \hspace{0.1in} \epsilon \sim \mathcal{N}(0, \sigma^2 I)
\end{gathered}
\end{equation}
The above smoothed classifier $S_c$ is provably robust against the $l_{2}$ perturbations with certified radius $R_c$ (refer to Sec.~\ref{sec:prelims}). For high performance on the certified defense, high $R_c$ is required, which is directly proportional to $\underline{p_1}$ (eq.~\ref{eq3}). The probability ($\underline{p_1}$) of predicting the most probable class ($c_1$) i.e. confidence depends on the performance of the denoiser network $D_n$. However, $D_n$ trained on the given limited training data $D_{train}^{lim}$ using the existing technique~\cite{salman2020denoised} yields poor certification (shown in Fig.~\ref{fig:motivation}). Even when we attempt to minimize the overfitting of $D_n$ on $D_{train}^{lim}$ ($\because$ $|D_{train}^{lim}| \ll |D_{train}|$) using different traditional approaches such as weight decay, augmentation, and mix-up strategies, we observe only a minor improvement (refer to Tables~\ref{tab:table1},~\ref{tab:table2}) in certified accuracy. Hence, we propose our method (`DE-CROP') that overcomes this limitation by crafting boundary and interpolated samples using $D_{train}^{lim}$,  followed by using them to train the denoiser with appropriate losses.
\begin{figure}[htp]
\centering
\centerline{\includegraphics[width=0.49\textwidth]{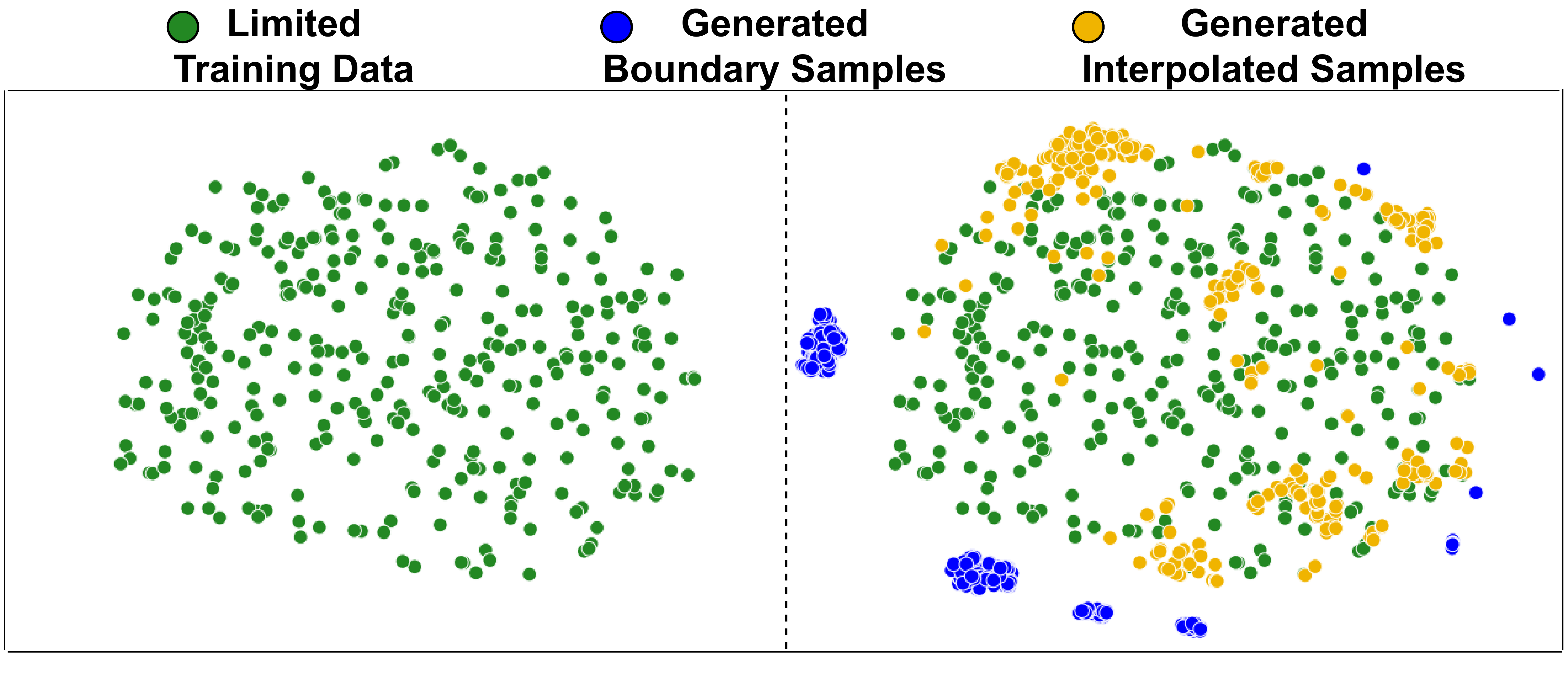}}
\caption{t-SNE visualization of the pretrained base classifier's features for training samples of a particular class of CIFAR-$10$. Our generated samples (interpolated and boundary) increases the feature diversity of limited-original training data.   
}
\label{fig:tsne}
\end{figure}
\begin{figure*}[htp]
\centering
\centerline{\includegraphics[width=\textwidth]{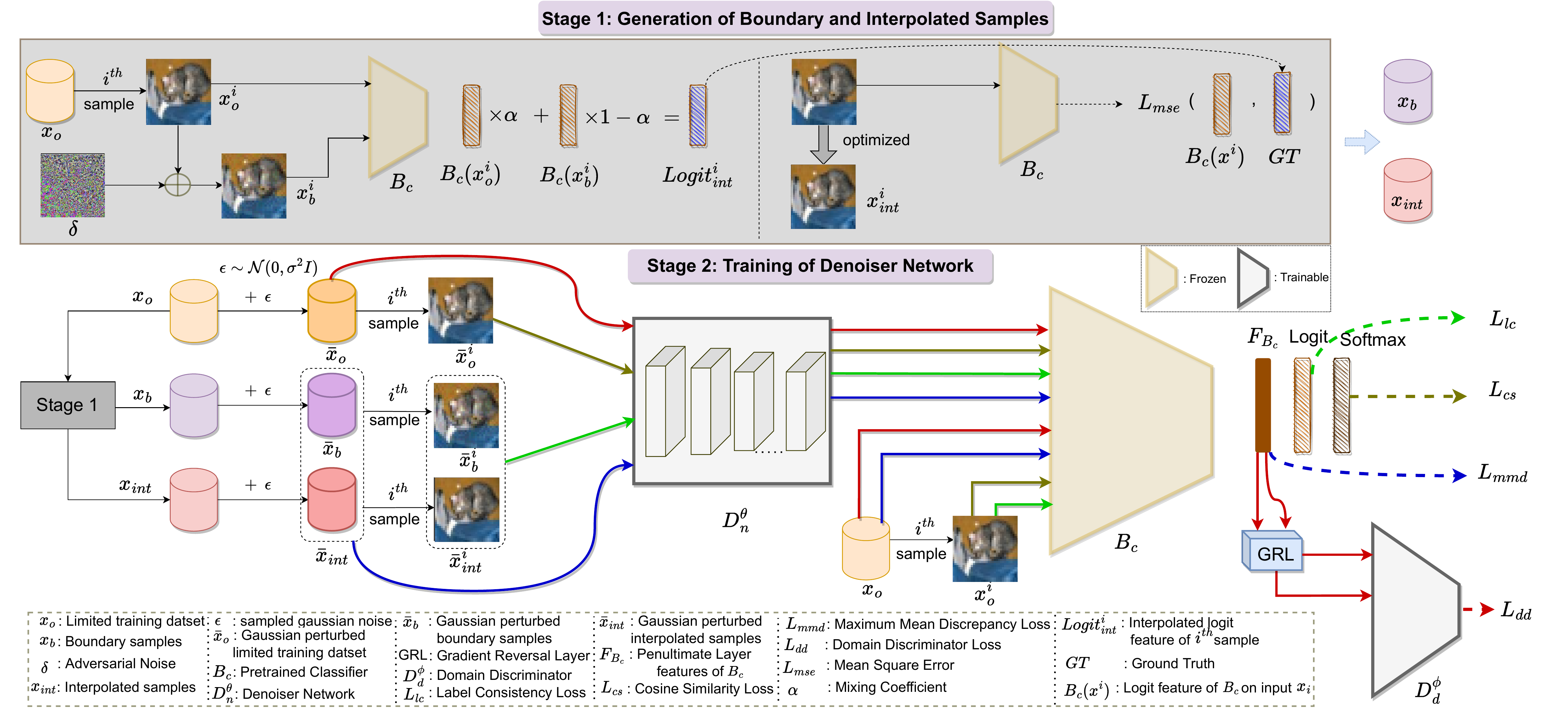}}
\caption{Different stages involved in the proposed method (\textit{DE-CROP}) that provides robustness guarantees on pretrained classifier ($B_c$) against $l_2$ perturbations using only limited training samples ($x_o$). Samples crafted via adversarial attack act as proxy \textit{boundary} samples ($x_b$). Using $x_o$ and $x_b$, \textit{interpolated} samples ($x_{int}$) are generated in stage $1$. The generated samples along with limited training data are used in stage $2$ to train the denoiser network ($D_n^\theta$) by aligning the feature representation at instance and distribution levels using $L_{lc}$, $L_{cs}$, $L_{mmd}$ and $L_{dd}$ losses. The forward pass to compute these losses are shown by different colors.
}
\label{fig:approach}
\end{figure*}
\vspace{-0.15in}
\subsection{Generation of Boundary \& Interpolated sample}
\label{subsec:generation}
In Fig.~\ref{fig:tsne}, we show the visualization of the logit layer features of the pretrained base classifier $B_c$ on a particular class via t-SNE plot. The class-features corresponding to 
available limited class samples of $D_{train}^{lim}$ is shown in 
green color (on left). 
We focus on improving the feature diversity of the limited training samples. For this, we first estimate class-boundary samples that would have respective features in the boundary region of the t-SNE. 

Adversarial attacks~\cite{madry2017towards} synthesize samples via an optimization procedure that carefully perturbs the input sample with a small human-imperceptible noise (i.e., adversarial noise). The model gets fooled on such samples (i.e., adversarial samples) as the prediction gets flipped to some other class. As these samples cross the decision boundary, which are constructed by adding a small noise to the input original sample, they often lie very close to the decision boundary. Moreover, they are human-imperceptible and preserve the class semantics of the input class sample. Hence, adversarial samples serve as a good candidate for a proxy of class-boundary samples. For any $i^{th}$ training sample of $D_{train}^{lim}$ (i.e., $x_o^i$), we obtain boundary sample ($x_b^i$) by computing adversarial noise $\delta$ for which the following relation holds:
\begin{equation}
\begin{gathered}
x_b^i=x_o^i+\delta, \\\left\lVert\delta\right\rVert_\infty<\epsilon, \hspace{0.1in}\epsilon>0, \hspace{0.1in}label(B_c(x_b^i)) \ne label(B_c(x_o^i)) 
\end{gathered}
\end{equation}
Next, we obtain interpolated features by performing a mixup between the features of generated boundary sample $x_b^i$ and original training sample $x_o^i$ as follows:
\begin{equation}
\begin{gathered}
Logit_{int}^i = \alpha \times B_c (x_o^i) + (1-\alpha) \times B_c (x_b^i) 
\end{gathered}
\end{equation}
Here $\alpha$ is the mixing coefficient. The interpolated features, the class-boundary features, and the original limited class-sample features are highlighted in yellow, blue, and green colors respectively, in the t-SNE plot (Fig.~\ref{fig:tsne}). These interpolated features being label preserving help to improve the feature-diversity 
leading to improved certified accuracy (refer to Sec.~\ref{gen_data}). We craft the input sample corresponding to the interpolated feature $Logit_{int}^i$ (i.e., $x_{int}^i$) by perturbing the original sample $x_o^i$ to match its feature response to $Logit_{int}^{i}$ using mean square error loss ($L_{mse}$). Mathematically, we obtain interpolated sample $x_{int}^i$ as follows:
\begin{equation}
\label{eq7}
x_{int}^i \leftarrow \underset{x}{\mathrm{min}} \hspace{0.1in} L_{mse}(B_c(x), Logit_{int}^i)
\end{equation}
where $x$ is initialized with $x_o^i$ and kept trainable with ground truth as $Logit_{int}^i$. The model $B_c$ is non-trainable, but the gradients are allowed to backpropagate from the model to update the input $x$. Thus, we obtain the boundary and the interpolated samples corresponding to each training sample. 
                
Refer supplementary for visualization of both the boundary and interpolated samples in the input space. All of them preserve the class semantics. Our generated boundary samples $x_b$ and interpolated samples $x_{int}$ are used along with limited training samples $x_o$ to train the denoiser network $D_n$, which we discuss in the subsequent subsection.
\subsection{Training of the denoiser network ($D_n$)}
\label{subsec:denoiser_training}
The denoiser network $D_n$ is attached before the pretrained base classifier $B_c$ to make it suitable for randomized smoothing. Apart from this, we also add a domain discriminator network $D_d$ whose input is the normalized features of the penultimate layer of $B_c$. The discriminator $D_d$ learns to distinguish between distribution of clean samples and distribution of denoised outputs of gaussian perturbed input samples. 
Motivated from domain adaptation literature~\cite{ganin2016domain}, we use gradient reversal layer (GRL) before feeding the normalized penultimate layer features to the discriminator that allows normal forward pass but reverses the direction of gradient in the backward pass. As a consequence, this negative gradients backpropagates to the denoiser network $D_n$ that helps it to produce denoised output which yields indistinguishable domain-invariant features on the pretrained $B_c$ classifier. 

Refer to supplementary for architectural details of the networks $D_n$ and $D_d$. The overall steps involved in the proposed framework (`DE-CROP’) are also shown in Fig.~\ref{fig:approach}. The network is trained with our generated data and limited training data $D_{train}^{lim}$ by using different losses aimed at different objectives - label consistency to ensure correct predictions on $B_c$ and matching of high-level feature similarity obtained on $B_c$ for denoised output of gaussian perturbed input and clean original training samples both at the sample level and distribution level. The respective losses are described below:\\
\textbf{Ensuring label consistency}:
Similar to~\cite{salman2020denoised}, we use cross entropy loss ($L_{ce}$) to ensure that the label predicted by the pretrained network $B_c$ on original clean data and denoised output of its gaussian perturbed counterpart are same.
\begin{equation}
\begin{split}
L_{lc} = 1/N^k\sum\nolimits_{i=1}^{N^k} L_{ce}(softmax(B_c(D_n(\bar{x}_o^i))),\\ label(B_c(x_o^i)))
\end{split}
\end{equation}
\textbf{Enforcing feature similarity at sample level}: The pretrained base classifier $B_c$ trained on complete training data $D_{train}$ is highly discriminative. To leverage this on our crafted data, we use cosine similarity loss at the logits of the $B_c$ network to encourage logit features on the denoised output of our generated data to be as discriminative as the features of the limited original training samples $D_{train}^{lim}$ by maximizing this loss. 
\begin{equation}
\begin{gathered}
L_{cs} = 1/N^{k}   \sum\nolimits_{i=1}^{N^k} (\hspace{0.05in}CS(B_c(D_n(\bar{x}_b^i)), B_c(x_o^i)) +\\ 
                                          CS(B_c(D_n(\bar{x}_{int}^i)), B_c(x_o^i))\hspace{0.05in})\text{,}
\hspace{0.1in} \text{s.t.} \hspace{0.05in}CS(w, z) = \frac{w^{T}z}{\norm{w}\norm{z}}
\end{gathered}
\end{equation}
\textbf{Enforcing feature similarity at distribution level}: Unlike $L_{cs}$ which is applied at sample level, here we enforce distribution level matching between the set of our denoised generated data and the set of limited original training data by using maximum mean discrepancy (MMD)~\cite{gretton2012kernel} loss on the normalized pre-logit layer of $B_c$ network.  
\begin{equation}
\begin{gathered}
L_{mmd} = MMD(F_{B_c}(D_n(\bar{x}_b)),F_{B_c}(x_o)) +\\ MMD(F_{B_c}(D_n(\bar{x}_{int})),F_{B_c}( x_o))
\end{gathered}
\end{equation}
Moreover, we also train the domain discriminator network $D_d$ (parameterized by $\phi$) using binary cross entropy loss ($L_{bce}$) to distinguish the distribution of gaussian perturbed samples and clean samples. 
\begin{equation}
\begin{gathered}
L_{dd} = \sum_{x^i \in D_{train}^{lim}\cup \hspace{0.01in}D_n(\bar{D}_{train}^{lim})}  L_{bce}(D_d(F_{B_c}(x^i)),d^i)
\end{gathered}
\end{equation}
Here, if $x^i \in D_{train}^{lim}$ then $d^i=1$ and if $x^i \in D_n(\bar{D}_{train}^{lim})$ then $d^i=0$ . The negative gradients are backpropagated via GRL~\cite{ganin2015unsupervised} (multiplies calculated gradient by $-1$), to update the parameters $\theta$ of denoiser network $D_n$ such that features of limited training data $D_{train}^{lim}$ and its corresponding denoised output of gaussian corrupted data ($D_n(\bar{D}_{train}^{lim})$) on network $B_c$ are domain invariant.

Hence, the total loss can be written as follows:
\begin{equation}
\label{eq12}
L(D_n^\theta, D_d^\phi) =   \beta_1 L_{lc} - \beta_2 L_{cs} +  \beta_3 L_{mmd} + \beta_4 L_{dd}
\end{equation}
At test time, the trained denoiser network $D_n$ with optimal parameters $\theta^{*}$ prepended with base classifier $B_c$ is used for evaluation. 
\section{Experiments}
\label{sec:experiments}
We demonstrate the effectiveness of our proposed approach (DE-CROP) on two widely-popular image-classification datasets, namely, CIFAR-10 \cite{krizhevsky2009learning} and Mini-ImageNet \cite{vinyals2016matching}. We limit the training set size of the datasets mentioned above by randomly selecting $1\%$ and $10\%$ samples respectively (ensuring class balance) from the $D_{train}$. Our baseline corresponds to training the denoiser using the $L_{lc}$ loss (similar to~\cite{salman2020denoised}). We fix the selected samples and use a ResNet-$110$ and ResNet-$12$ \cite{he2016deep} network (for CIFAR-$10$ and Mini-ImageNet respectively) as our pre-trained classifiers ($B_c$) with a value of $\sigma$ as $0.25$ for all our ablations and state-of-the-art comparisons, unless otherwise specified. 
Refer to supplementary for additional ablations on different values of noise strength $\sigma$ ($0.12$, $0.50$, $1.00$), quantity of limited training data $D_{train}^{lim}$ ($5\%$, $10\%$, $20\%$, $100\%$) and choice of architecture for the pretrained classifier $B_c$. 
We set the weights for the final loss-equation (refer eq.~\ref{eq12}) i.e. $\beta_1$, $\beta_2$, $\beta_3$, $\beta_4$ as $1$, $4$, $4$, $1$ respectively. In the subsequent subsections, we first show limited benefit of conventional techniques, followed by benefits of each component in DE-CROP and a comparison with state-of-the-art techniques.

\subsection{Improving Certification on Limited Training Data via Conventional Techniques}
$D_{n}$ trained with the $L_{lc}$ objective proposed by \cite{salman2020denoised} severely overfits in the presence of limited data resulting in poor certification accuracy (refer Fig.~\ref{fig:motivation}). In this section, we explore whether conventional supervised-learning techniques such as explicit regularization and data-augmentation are equipped to meaningfully improve certification accuracy when dealing with limited data.

\begin{table}[htp]
\centering
\scalebox{0.8}{
\begin{tabular}{|c|c|ccc|}
\hline
\multirow{2}{*}{\textbf{Method}} & \textbf{\begin{tabular}[c]{@{}c@{}}Standard\\ Certified\end{tabular}} & \multicolumn{3}{c|}{\textbf{Robust Certified}} \\ \cline{2-5} 
             & (r=0.00) & \multicolumn{1}{c|}{(r=0.25)} & \multicolumn{1}{c|}{(r=0.50)} & (r=0.75) \\ \hline
Without Reg. & 20.60    & \multicolumn{1}{c|}{4.60}    & \multicolumn{1}{c|}{0.80}    & 0.00      \\ \hline
$L_{1}$ Reg.      & 22.60    & \multicolumn{1}{c|}{5.80}    & \multicolumn{1}{c|}{0.60}    & 0.00      \\ \hline
$L_{2}$ Reg.      & \textbf{27.80}    & \multicolumn{1}{c|}{\textbf{7.80}}    & \multicolumn{1}{c|}{\textbf{0.80}}    & 0.00      \\ \hline
\end{tabular}
}
\caption{Effect of adding weight decay regularizer ($L_{1}$ and $L_{2}$) on limited data certification against adversarial attacks. $L_{2}$ regularization obtains better certified standard and robust accuracies.}
\label{tab:table1}
\end{table}
In Table~\ref{tab:table1}, we regularize the $D_{n}$ by applying $L_{1}$ and $L_{2}$ reg. (regularization) \cite{hanson1988comparing}. We observe that $L_{2}$ reg. improves certified 
performance, whereas $L_{1}$ reg. only results in a marginal improvement. Although $L_{2}$ reg. limits the overfitting effect in the presence of limited data, it can’t help in improving the inherent data diversity. Thus, in Table~\ref{tab:table2}, we explore whether traditional affine and specialized augmentation methods (such as mixup \cite{zhang2017mixup} and cutmix \cite{yun2019cutmix}), when combined with $L_{2}$ reg. further boost performance. 

Rows $2$-$4$ in Table~\ref{tab:table2} correspond to affine-transformations where the intensity of the augmentation is increased progressively (i.e. policy$1$$<$policy$2$$<$policy$3$). We also experiment with widely-popular augmentation techniques: ‘mixup’ and 'cutmix'. Surprisingly, we observe that policy $1$(row-$2$; the lightest augmentation) performs the best, followed closely by ‘no-aug.’. We hypothesize that if the augmentation-policy is too aggressive, $B_c$ can make incorrect predictions leading to noisy gradients for the $D_n$ to learn from, resulting in poor certification accuracy.

\begin{table}[htp]
\centering
\scalebox{0.8}{
\begin{tabular}{|c|c|ccc|}
\hline
\multirow{2}{*}{\textbf{Method}} & \textbf{\begin{tabular}[c]{@{}c@{}}Standard\\ Certified\end{tabular}} & \multicolumn{3}{c|}{\textbf{Robust Certified}} \\ \cline{2-5} 
                                                          & (r=0.00) & \multicolumn{1}{c|}{(r=0.25)} & \multicolumn{1}{c|}{(r=0.50)} & (r=0.75) \\ \hline
No Aug.                                                   & 27.80    & \multicolumn{1}{c|}{7.80}    & \multicolumn{1}{c|}{0.80}    & 0.00      \\ \hline
\begin{tabular}[c]{@{}c@{}}Aug.\\ (policy 1)\end{tabular} & \textbf{29.80}    & \multicolumn{1}{c|}{\textbf{9.20}}    & \multicolumn{1}{c|}{\textbf{1.40}}    & 0.00      \\ \hline
\begin{tabular}[c]{@{}c@{}}Aug.\\ (policy 2)\end{tabular} & 26.40    & \multicolumn{1}{c|}{7.40}    & \multicolumn{1}{c|}{1.00}    & 0.00      \\ \hline
\begin{tabular}[c]{@{}c@{}}Aug\\ (policy 3)\end{tabular}  & 21.00     & \multicolumn{1}{c|}{3.00}     & \multicolumn{1}{c|}{0.20}    & 0.00      \\ \hline
Mixup                                                     & 24.20    & \multicolumn{1}{c|}{5.40}    & \multicolumn{1}{c|}{0.60}    & 0.00      \\ \hline
Cutmix                                                    & 19.80    & \multicolumn{1}{c|}{3.20}    & \multicolumn{1}{c|}{0.00}    & 0.00      \\ \hline
\end{tabular}
}
\caption{Investigating the effect of augmentation at different intensity levels (policies), mixup and cutmix strategies in minimizing overfitting on limited training data. 
The augmentation with light intesnity (policy 1) yields marginal improvement against l2 perturbations of different radii compared to no-augmentation strategy.}
\label{tab:table2}
\end{table}

Thus, a combination of policy $1$ and L$2$-reg. improves the certified standard and robust accuracies. We use this combination as a baseline for further experiments upon which we make orthogonal improvements. 

\subsection{Effectiveness of our Generated Data}
\label{gen_data}
One of the critical reasons for the drop in certified accuracy is the lack of diversity in the limited data. We address this problem by generating synthetic samples that provide diversity in the feature space. As elaborated in the proposed section (refer Sec.~\ref{subsec:generation}): adversarial samples (termed as boundary-samples i.e. $x_b$) serve as a good candidate for this task since they flip the classifier prediction with minimum possible perturbations, thus allowing the $D_n$ to train on samples from less-dense boundary regions. Moreover, we also generate samples whose features are an interpolation between the original and boundary samples, crafted by minimizing the $L_{mse}$ loss  between the interpolated logits and the generated sample logits (as described by eq.~\ref{eq7}). 

We empirically validate our motivation in Table~\ref{tab:table3}, where we observe that using $L_{cs}$ on both $x_{b}^{i}$ and corresponding $x_{int}^{i}$ leads to a massive improvement in performance over the baseline. Interestingly, the gain in performance when using only $x_{b}^{i}$ is comparatively modest. This observation further reinforces our intuition regarding the complementary nature of the information provided by the $x_{b}^{i}$ and $x_{int}^{i}$. Moreover, contrary to adversarial training that generates adversarial samples at every iteration during the training time, we only need to generate the boundary and interpolated samples once (amounting to negligible increase in training time), as pre-trained classifier $B_c$ is fixed and not trainable.
\vspace{-0.15in}
\begin{table}[htp]
\centering
\scalebox{0.8}{
\begin{tabular}{|c|c|ccc|}
\hline
\multirow{2}{*}{\textbf{Method}} &
  \textbf{\begin{tabular}[c]{@{}c@{}}Standard \\ Certified\end{tabular}} &
  \multicolumn{3}{c|}{\textbf{Robust Certified}} \\ \cline{2-5} 
         & (r=0.00) & \multicolumn{1}{c|}{(r=0.25)} & \multicolumn{1}{c|}{(r=0.50)} & (r=0.75) \\ \hline
Baseline & 29.80    & \multicolumn{1}{c|}{9.20}    & \multicolumn{1}{c|}{1.40}    & 0.00      \\ \hline
\begin{tabular}[c]{@{}c@{}}Ours (with boundary\\ samples)\end{tabular} &
  31.60 &
  \multicolumn{1}{c|}{7.80} &
  \multicolumn{1}{c|}{1.00} &
  0.20 \\ \hline
\begin{tabular}[c]{@{}c@{}}Ours (with boundary +\\ interpolated samples)\end{tabular} &
  \textbf{48.80} &
  \multicolumn{1}{c|}{\textbf{22.00}} &
  \multicolumn{1}{c|}{\textbf{6.00}} &
  \textbf{0.80} \\ \hline
\end{tabular}
}
\caption{Benefit of our generated samples in improving certification. Using boundary and interpolated samples, we obtain significant boost in certified accuracy on original and l2 perturbed data.}
\label{tab:table3}
\end{table}
\vspace{-0.15in}
\subsection{Distribution Alignment: Enhancing Certification in Limited Data Setup}
\label{dist_align}

In Table~\ref{tab:table4}, we observe that using domain-discriminator ($D_d$) along with the previously introduced $L_{cs}$ loss works quite well as the standard certified accuracy improves by ~6$\%$ and certified robust accuracy improves by 4$\%$ (at r=$0.25$) compared to only using $L_{cs}$.

We also explore whether equipping the $D_n$ with explicit distribution discrepancy losses such as $L_{mmd}$ would also work well in combination with the $D_d$. Intuitively, applying $L_{mmd}$ along with $L_{cs}$ should make the job of $D_d$ harder, leading to a better $D_d$. Consequently, improving $B_c$’s feature representation via the negative gradients of domain-discriminator loss ($L_{dd}$). 

We indeed observe this in Table~\ref{tab:table4}, where using $L_{mmd}$ in combination with the $L_{cs}$ and the $L_{dd}$ setup leads to an improvement in performance across both standard and robust accuracies (across radii). Thus, using $L_{mmd}$ and $L_{dd}$ in conjunction with the previously discussed $L_{cs}$ and $L_{lc}$ constitutes our final approach: DE-CROP
\begin{table}[htp]
\centering
\scalebox{0.8}{
\begin{tabular}{|c|c|ccc|}
\hline
\multirow{2}{*}{\textbf{Method}} &
  \textbf{\begin{tabular}[c]{@{}c@{}}Standard\\ Certified\end{tabular}} &
  \multicolumn{3}{c|}{\textbf{Robust Certified}} \\ \cline{2-5} 
                      & (r=0.00) & \multicolumn{1}{c|}{(r=0.25)} & \multicolumn{1}{c|}{(r=0.50)} & (r=0.75) \\ \hline
Baseline              & 29.80    & \multicolumn{1}{c|}{9.20}    & \multicolumn{1}{c|}{1.40}    & 0.00      \\ \hline
Ours (instance level) & 48.80    & \multicolumn{1}{c|}{22.00}     & \multicolumn{1}{c|}{6.00}    & 0.80     \\ \hline
\begin{tabular}[c]{@{}c@{}}Ours (instance + \\ distribution level via \\ discriminator)\end{tabular} &
  54.00 &
  \multicolumn{1}{c|}{26.00} &
  \multicolumn{1}{c|}{6.80} &
  1.80 \\ \hline
\begin{tabular}[c]{@{}c@{}}Ours (instance + \\ distribution level via \\ discriminator and MMD)\end{tabular} &
  \textbf{57.60} &
  \multicolumn{1}{c|}{\textbf{27.20}} &
  \multicolumn{1}{c|}{\textbf{9.20}} &
  \textbf{2.20} \\ \hline
\end{tabular}
}
\caption{
Besides performance gains with instance-level feature matching, we observe further improvements in certified standard and robust accuracy when distributions of denoised and clean data are aligned in feature space via domain-discriminator and MMD.}
\label{tab:table4}
\end{table}
\subsection{Comparison with state-of-the-art}
In this section, we compare the effectiveness of our approach DE-CROP against 
state-of-the-art robustness certification techniques, namely, denoised-smoothing by Salman \etal~\cite{salman2020denoised} and gaussian-augmentation by Cohen \etal~\cite{cohen2019certified}.

Since Salman \etal~\cite{salman2020denoised} don’t report performance on limited data scenarios in their paper, we use the code from their official implementation\footnote{https://github.com/microsoft/denoised-smoothing} for evaluating their proposed method ($D_n$ with $L_{lc}$) in the presence of only $1\%$ (for CIFAR-$10$) and $10\%$ (for Mini-ImageNet) $D_{train}$. Similarly, for Cohen \etal~\cite{cohen2019certified}, we re-train the classifier with gaussian augmentation only on the available limited training data. Although Cohen \etal~\cite{cohen2019certified}’s technique is unfeasible for our problem setup as the API provider may not prefer re-training and replacing the deployed model, we still compare our performance as gaussian-augmentation often outperforms previous denoiser-based approaches in presence of full training data.

\begin{figure}[htp]
\centering
\centerline{\includegraphics[width=0.49\textwidth]{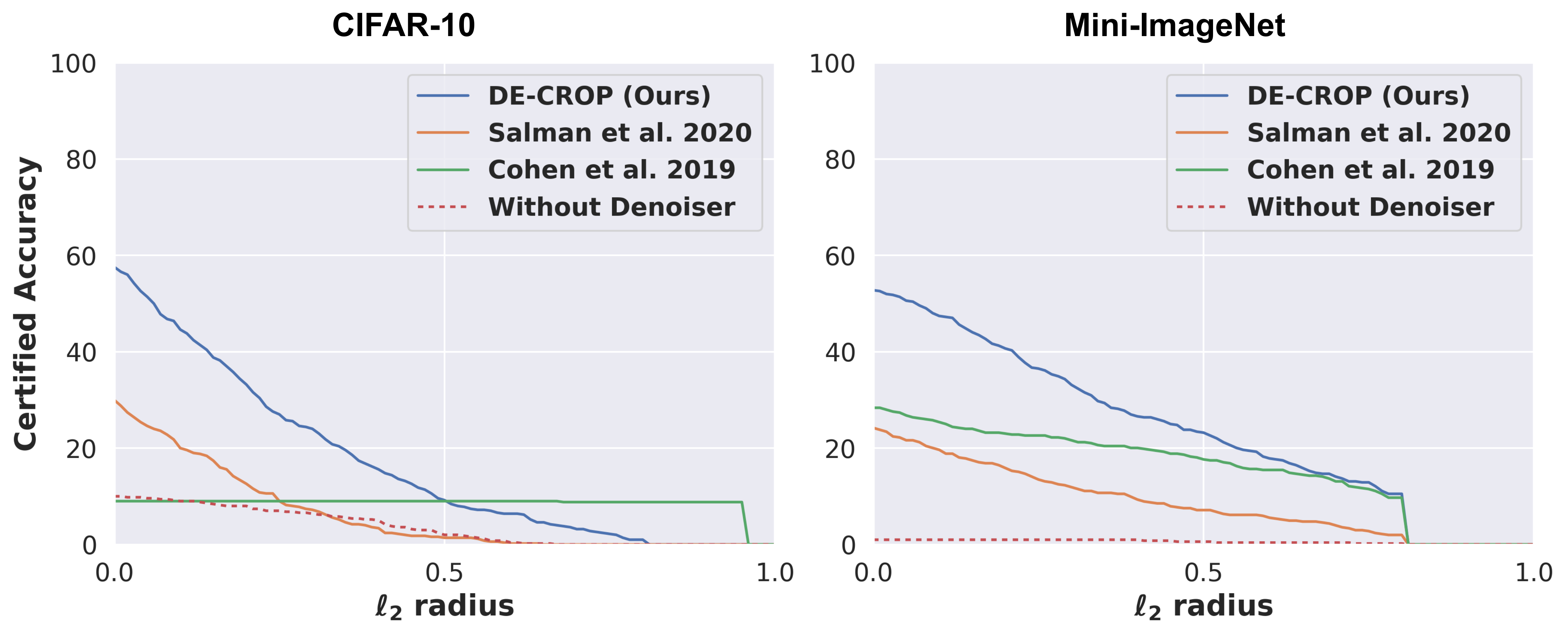}}
\caption{Performance comparison of our approach (\textit{DE-CROP}) with other methods. We comfortably outperform on both the datasets. We also compared with Cohen \etal where the classifier is trained from scratch unlike ours where retraining is not feasible. 
}
\label{fig:sota}
\end{figure}

As shown in Fig.~\ref{fig:sota}, our proposed approach DE-CROP comfortably outperforms Salman \etal~\cite{salman2020denoised} on CIFAR-$10$, improving the certified standard accuracy by $27\%$ and consistently improving robust accuracy across radii. The performance of Cohen \etal~\cite{cohen2019certified} drops to $10\%$ across all radii, indicating that the network behaves like a random baseline (i.e. predicting each class with equal probability irrespective of the input). We observe similar trends for Mini-ImageNet, where we comfortably outperform both Salman \etal~\cite{salman2020denoised} and Cohen \etal~\cite{cohen2019certified} further demonstrating the wide applicability of our approach. 
\vspace{-0.11in}
\subsection{Certification of Black Box Classifiers with \newline Limited Training Data}
In previous sections we assumed white-box access to the pre-trained base classifier $B_c$ i.e. we can backpropagate the gradients through $B_c$ to optimize the denoiser ($D_n$). However, this may not always be the case as the API provider can limit access to only $B_c$’s predictions (i.e. black-box) due to proprietary reasons. Since the black-box setup restricts the gradient information of $B_c$, we first use a black-box model stealing technique \cite{barbalau2020black} to train a surrogate model: $S_m$. We use $S_m$ which allows gradient backpropagation, to train $D_n$ using our proposed approach DE-CROP (refer Fig.~\ref{fig:approach}). Finally, for evaluation we use the denoiser (trained via $S_m$) to certify robustness of the black-box classifier $B_c$.

\begin{figure}[htp]
\centering
\centerline{\includegraphics[width=0.35\textwidth]{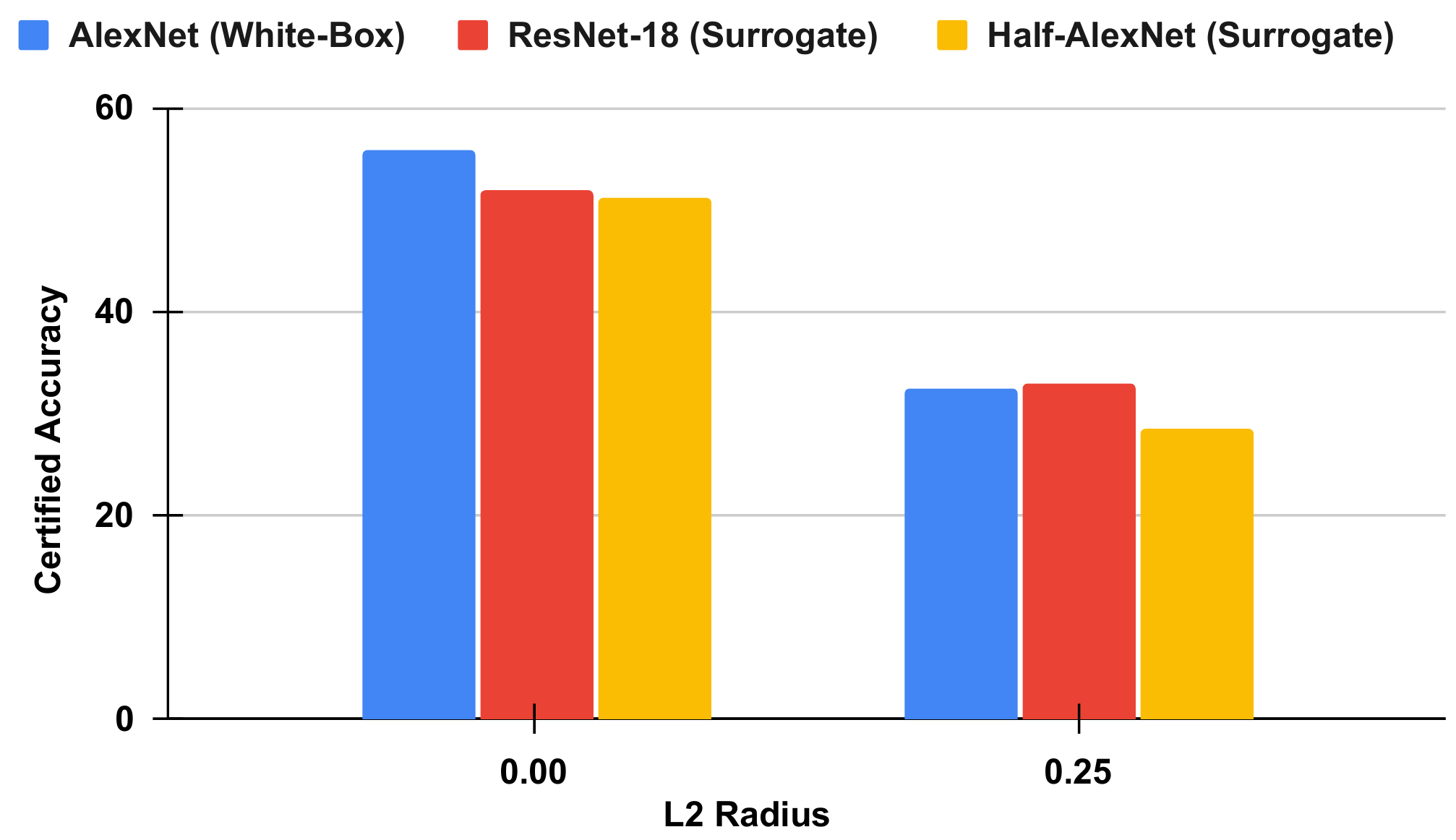}}
\caption{Investigating the efficacy of our method in black-box scenario (no access to pertrained model weights). We observe a minor drop in performance as compared to white-box setup for both certified standard accuracy ($l_2^r$=$0.00$) and robust accuracy ($l_2^r$=$0.25$).}
\label{fig:blackbox}
\end{figure}
In Fig.~\ref{fig:blackbox}, we compare the certification performance (of $B_c$) using denoisers trained via $S_m$ (`black-box access’) to the denoiser trained directly on $B_c$ (`white-box access’). We take $B_c$ as Alexnet and two different choices for $S_m$, namely ResNet-$18$ and Half-Alexnet. Our method DE-CROP yields very similar performance in the black box setting across different architectures of $S_m$. Also, the performance drop compared to white box setting is marginal, highlighting the suitability of our technique even when the pretrained classifier weights are not shared.
\section{Conclusion}
We presented our approach (DE-CROP), which for the first time, solves the problem of providing provable robustness guarantees to a pretrained classifier in the challenging limited training data settings. Our method comprises a two-step process - a) generation of boundary and interpolated samples ensuring feature diversity and b) effectively utilizing the generated samples along with limited training samples for training the denoiser using the proposed losses to ensure feature similarity between the denoised output and clean data at two different granularities (instance level and distribution level). We validate the efficacy of the generated data as well as the contribution of individual losses by extensive ablations and experiments across CIFAR-$10$ and Mini-ImageNet datasets. Moreover, our method works quite well in the black-box setting as it provides similar certification performance compared to the white-box setup. 

\vspace{0.1in}
\noindent{\textbf{Acknowledgements}} This work is partially supported by a Young Scientist Research Award (Sanction no. 59/20/11/2020-BRNS) from DAE-BRNS, India.

{\small
\bibliographystyle{ieee_fullname}
\bibliography{references}
}

\newpage
\onecolumn
\begin{center}
    \Large{\textbf{{\textit{Supplementary for:} \\``DE-CROP: Data-efficient Certified Robustness for Pretrained Classifiers"}}}

\end{center}

\setcounter{section}{0}
\setcounter{table}{0}
\setcounter{figure}{0}
\setcounter{equation}{0}
\vspace{18pt}
\hrule
\vspace{18pt}

\section{Certification performance on different noise strengths (varying $\sigma$)}
\begin{figure}[htp]
\centering
\centerline{\includegraphics[width=\textwidth]{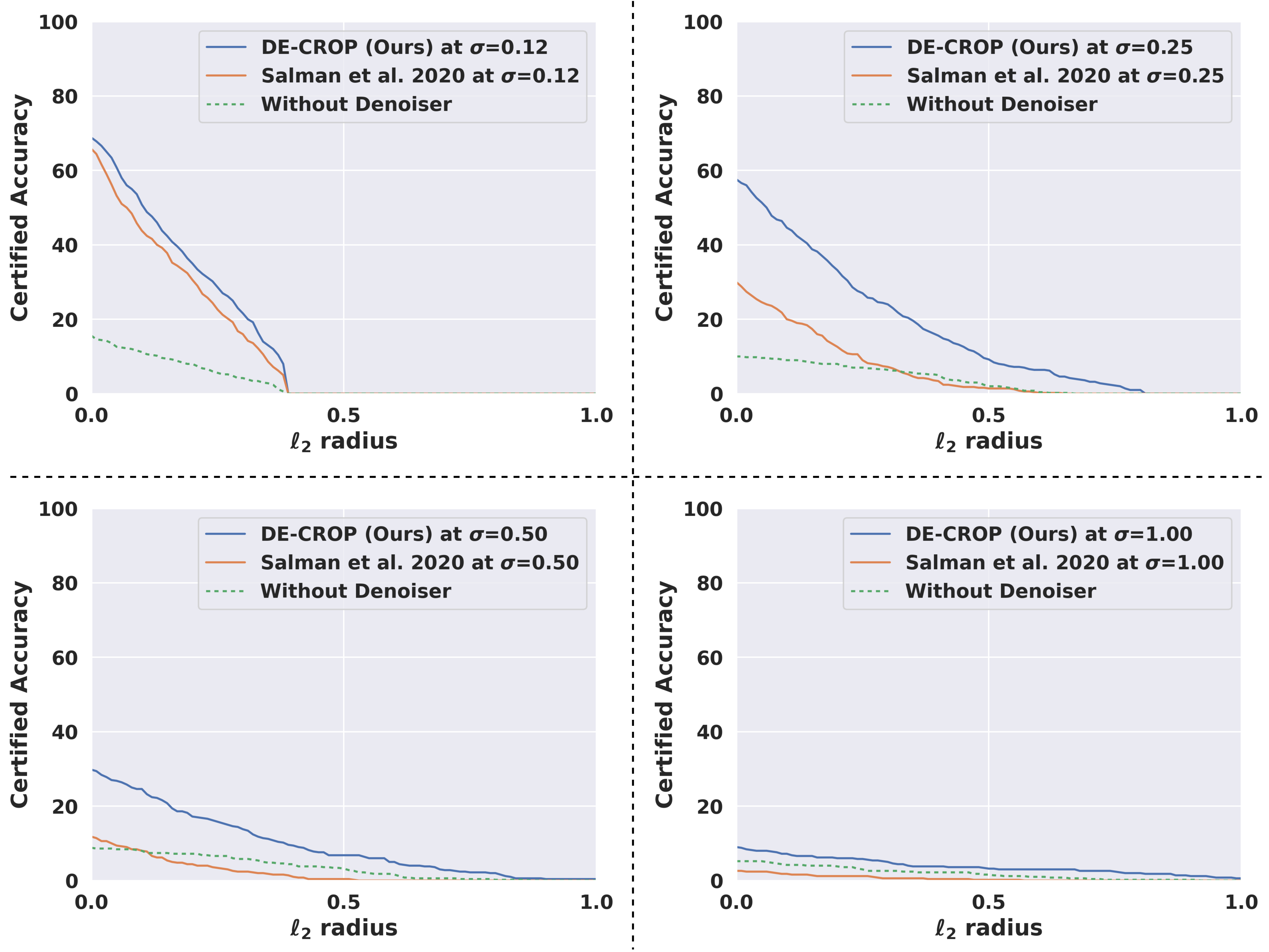}}
\caption{Our method (DE-CROP) consistently provides  better certified robustness across different noise strengths with $\sigma \in \{0.12, 0.25, 0.50, 1.00\}$. The results are presented on limited training data $D_{train}^{lim}$ which is $1\%$ of entire training samples $D_{train}$ of CIFAR-$10$. 
}
\label{fig:1}
\end{figure}
\newpage
\section{Certified guarantees on different quantity of limited training data}
\begin{figure}[htp]
\centering
\centerline{\includegraphics[width=\textwidth]{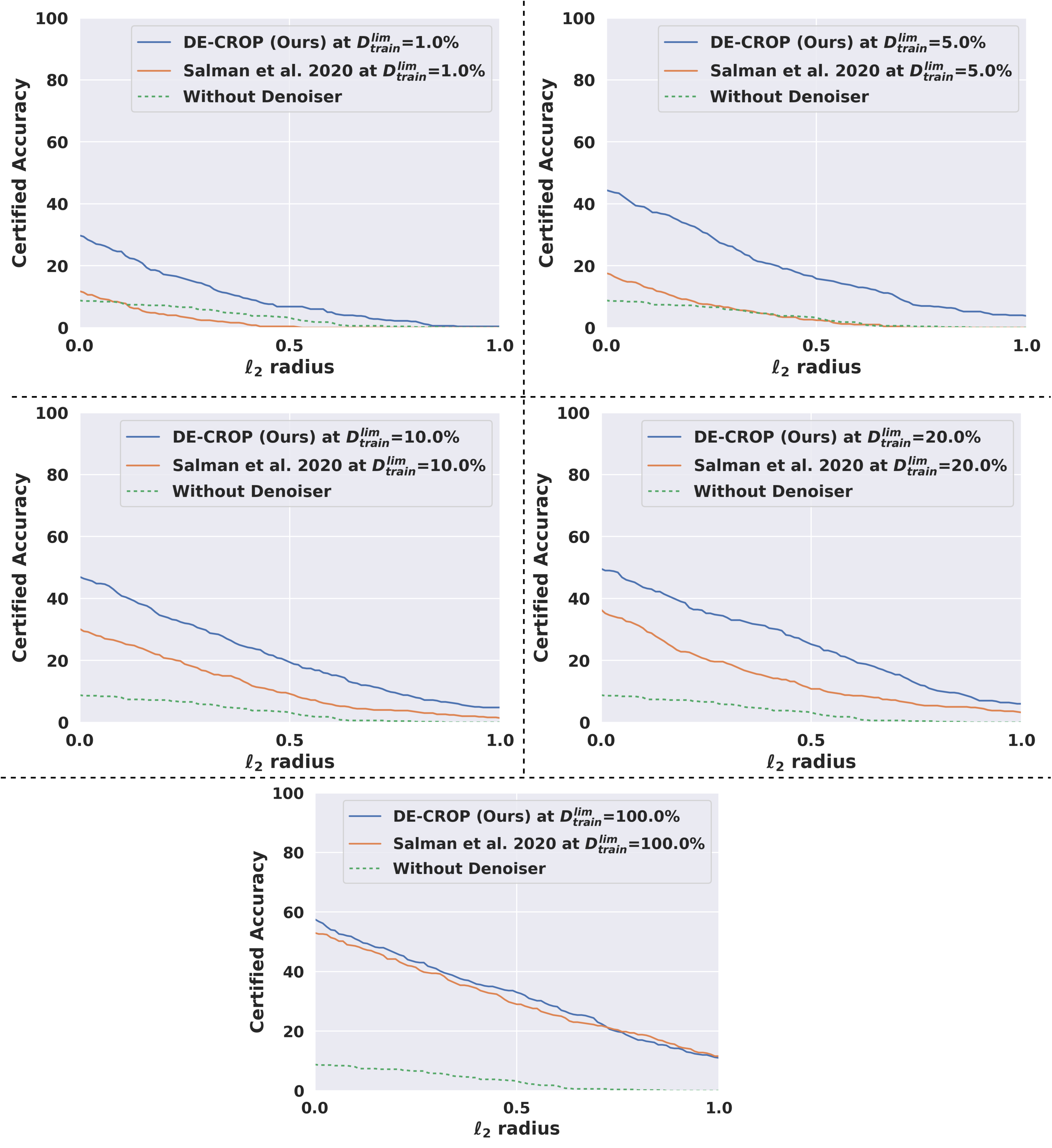}}
\caption{The amount of limited training data ($|D_{train}^{lim}|$) is $k\%$ of $D_{train}$. We perform ablation for different choices of $k$ (i.e. $1$, $5$, $10$, $20$ and $100$) and report our certified performance. DE-CROP significantly boosts the certified accuracy across the radii when the available training set size is small, showing the efficacy of our method for limited data settings. The benefit is limited when the amount of training samples is high. The experiments are conducted on CIFAR-$10$ dataset with noise strength $\sigma$ = $0.50$
}
\label{fig:2}
\end{figure}

\section{Sensitivity Analysis of mixing coefficient ($\alpha)$}
\begin{figure}[htp]
\centering
\centerline{\includegraphics[width=0.6\textwidth]{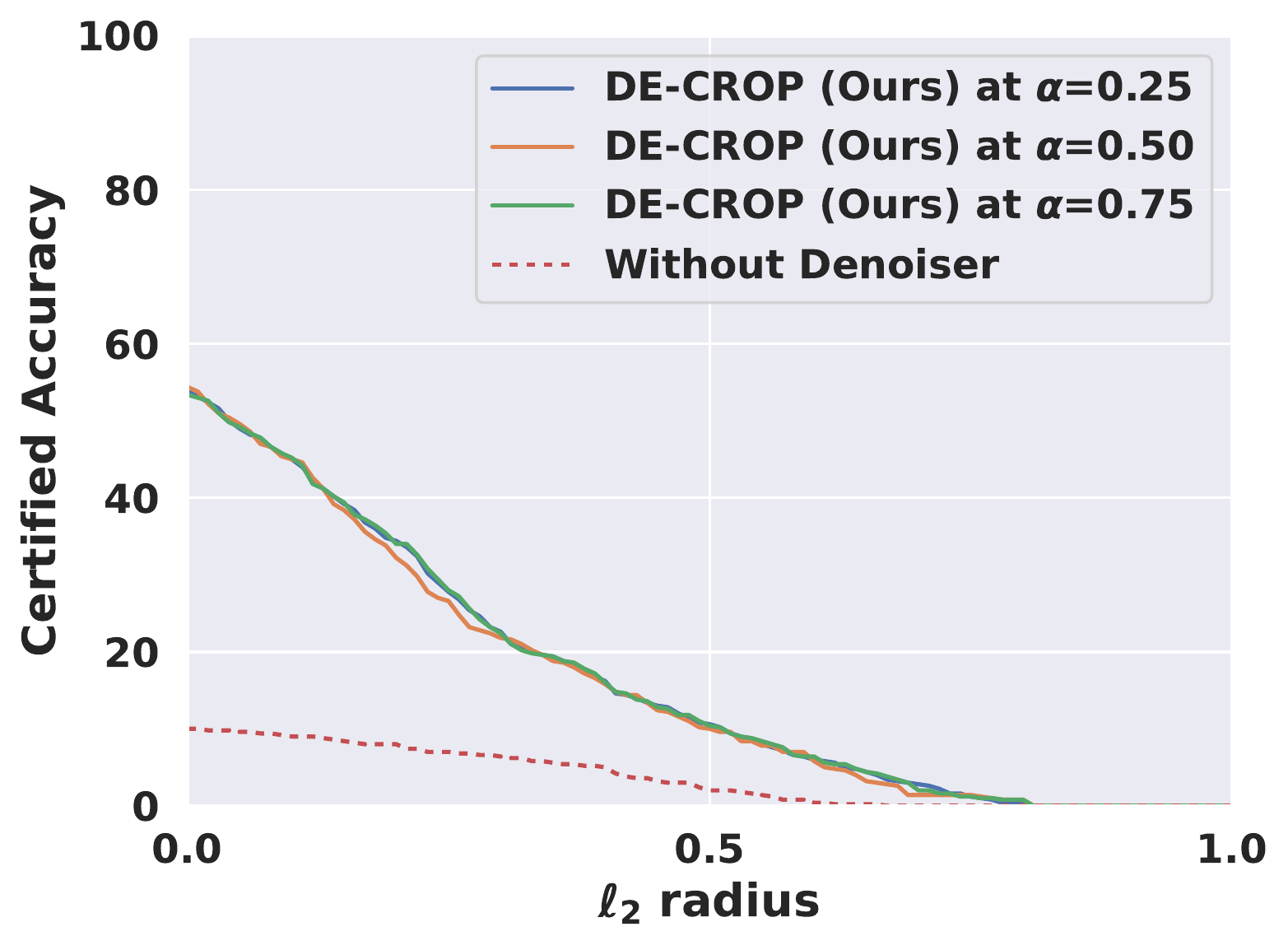}}
\caption{Ablation to determine the sensitivity of our proposed method (DE-CROP) to change in value of mixing coefficient ($\alpha$). We perform experiments with three different values of $\alpha$ (i.e. $0.25$, $0.50$, $0.75$) and observe that the certified performance of DE-CROP is stable across the broad range of $\alpha$ values allowing our technique to be easily adoptable without significant hyper-parameter tuning.
}
\label{fig:alpha_sensitivity}
\end{figure}

\section{Additional experimental details}
For all our experiments, we use three $1080$Ti $12$Gb cards. The additional details of specific components of our proposed approach (DE-CROP) are provided below:
\newline
\newline
\textbf{Training of base classifier ($B_c$)}: The base pretrained classifier $B_c$ is trained for $300$ epochs with cross-entropy loss and stochastic gradient descent optimizer. The initial learning rate is $0.1$, which reduces by a factor of $10$ every $100$th epoch.  
\newline
\newline
\textbf{Training of denoiser network ($D_n$)}: We use the architecture proposed by~\cite{zhangdncnn} for our denoiser ($D_n$). The network $D_n$ is a $17$ layers deep fully convolutional network that contains multiple blocks of convolutional and batch-norm layers followed by ReLU activation. We train the denoiser network $D_n$ for $600$ epochs, adopting the same learning routine as Salman \etal~\cite{salman2020denoised}. The Adam optimizer is used with a learning rate of $0.001$ for initial 50 epochs, followed by stochastic gradient descent optimizer with a learning rate of $0.001$ that reduces by a factor of $10$ every $200$th epoch. 
\newline
\newline
\textbf{Training of domain discriminator ($D_d$)}: We use a vanilla multi-layer perceptron architecture with one hidden layer of size $100$ units as the architecture for domain discriminator ($D_d$). While training the $D_d$, we initialize the $D_n$ by pre-training it using $L_{lc}$, $L_{cs}$ and $L_{mmd}$ losses. Then optimize the parameters of $D_d$ ($\phi$) using Adam optimizer with a learning rate of 1e-5.  Moreover, in order to ignore the noisy gradients from $D_d$ during the initial phase of training the domain discriminator we use a scheduler (similar to \cite{ganin2015unsupervised}) to weight that value of negative gradient by $\beta$ that is gradually increased over the course of training. We use the standard parameters for certification described by Salman \etal \cite{salman2020denoised}.
\newline
\newline
\textbf{Details on augmentations}: In the main draft (Table $2$ of Sec. $5.1$), we showed experimental results to demonstrate the effect of various augmentation policies in improving certification performance. In policy $1$, we apply the different transformations - ‘randomized cropping‘ followed by ‘random horizontal flipping with probability of $0.5$.  For policy $2$ we randomly select one of the following augmentations: ‘random horizontal flip (probability=$0.5$)’, ‘random vertical flip (with probability=$0.5$)’, ‘randomized cropping’, ‘random affine transform to translate (with parameters = ($0.25$, $0.50$) and rotate (degrees=$10$) the image. Whereas for policy $3$  all the augmentations of policy $2$ are applied in a sequential manner making it a stronger augmentation via composition of different transformations that are applied on a given sample.

\newpage
\section{Visualizations of Original and Generated Samples}
\begin{figure}[htp]
\centering
\centerline{\includegraphics[width=0.65\textwidth]{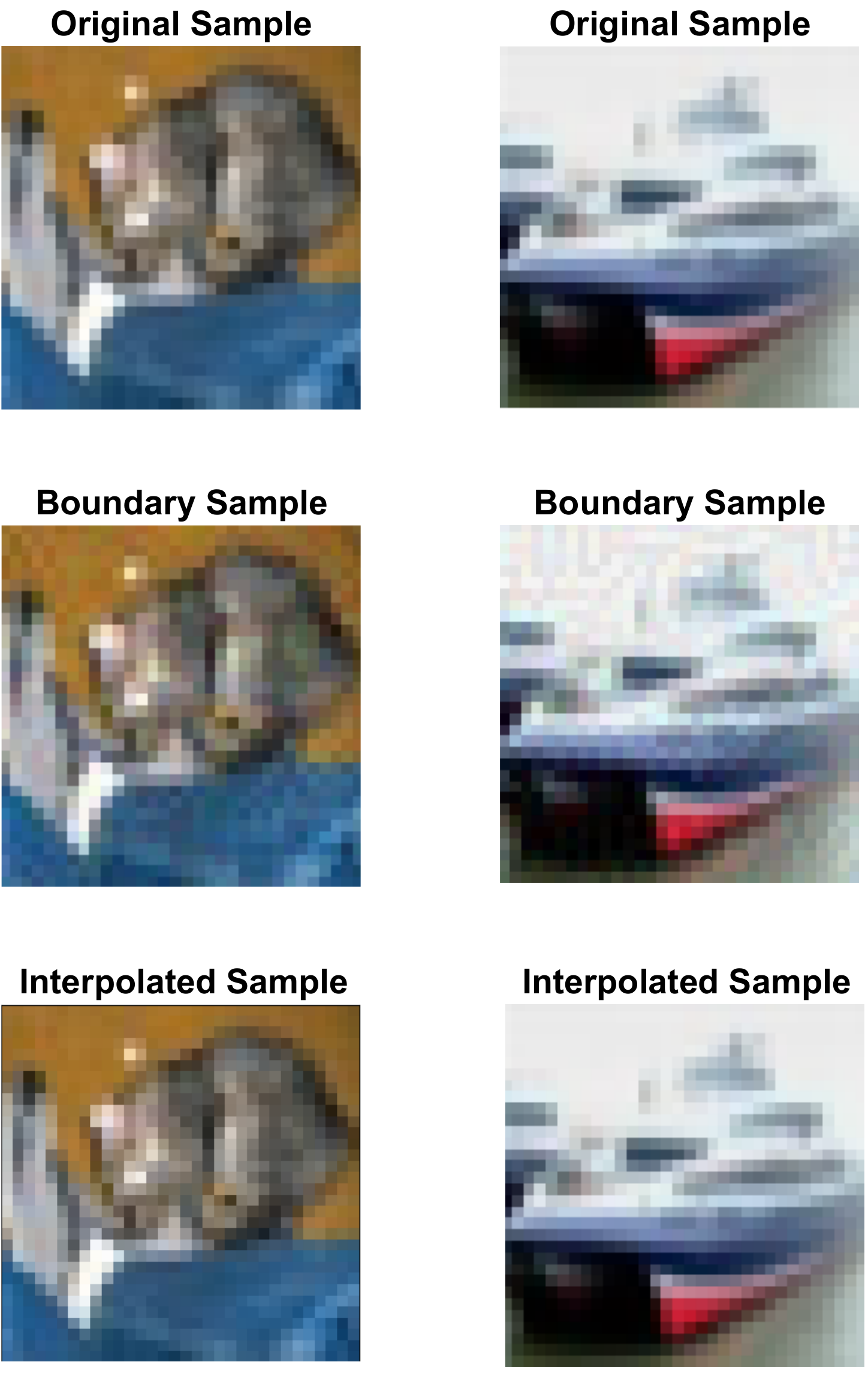}}
\caption{Visualization of the original, generated boundary and generated interpolated samples for CIFAR-$10$ dataset. The boundary samples are generated by perturbing the original sample with adversarial noise (refer eq. $5$ in the main draft) and interpolated samples are generated by minimizing the Mean Squared Error with interpolation of original and boundary samples logits as the ground truth (refer eq.~$7$ in the main draft). Our simple and intuitive sample generation technique avoids complicated and time consuming methods like generative adversarial networks. Moreover, the generated samples ensure preservation of class semantics while providing diversity in the feature space of pre-trained classifiers (ref Fig. $2$ in the main draft)}
\label{fig:3}
\end{figure}
\newpage
\section{Certifying pretrained models of different architecture using limited training data}
\begin{figure}[htp]
\centering
\centerline{\includegraphics[width=0.7\textwidth]{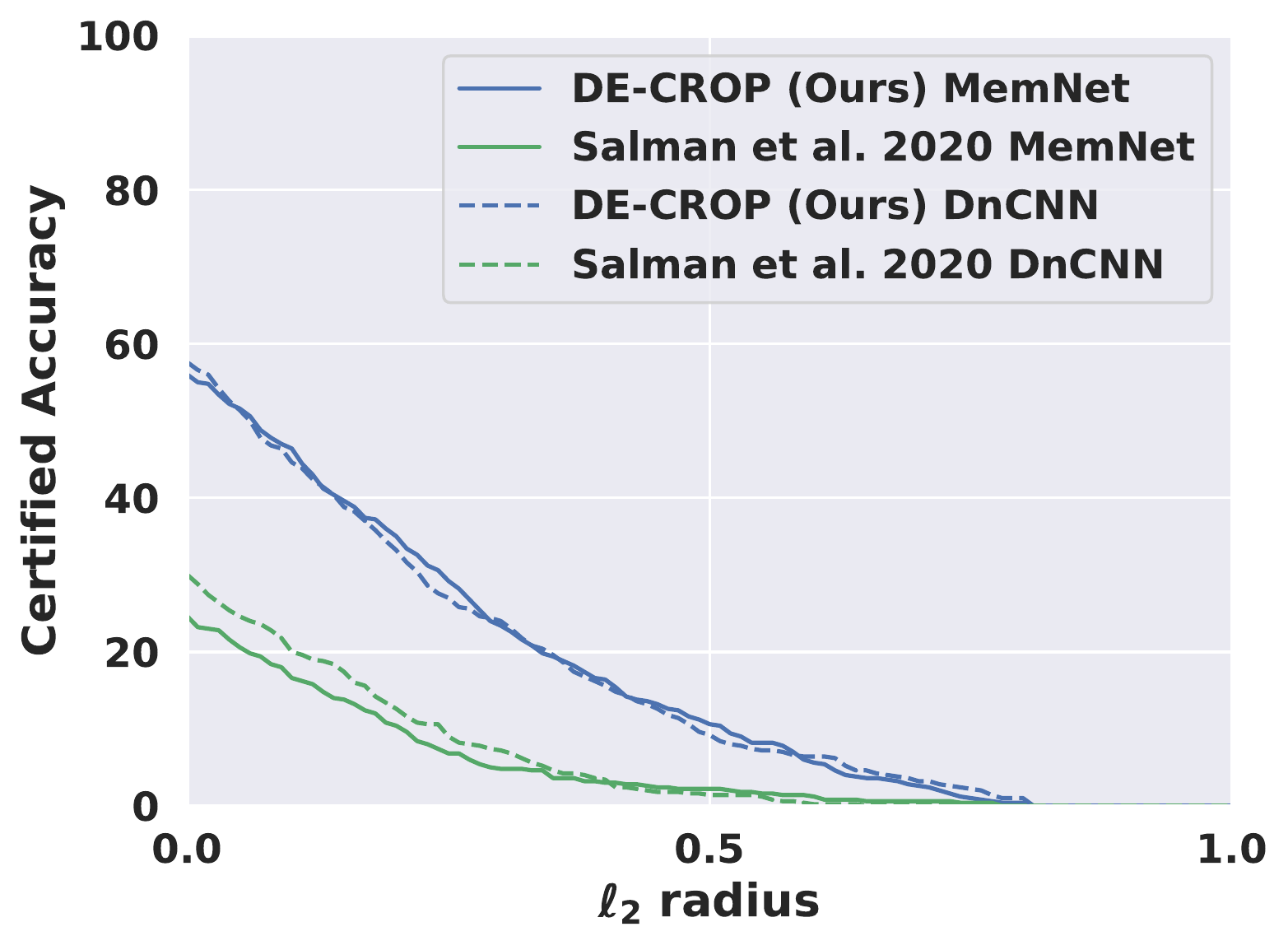}}
\caption{Performance comparison of our approach (DE-CROP) against Salman \etal \cite{salman2020denoised} across multiple choices of denoiser architectures i.e. DnCNN and MemNet. DE-CROP achieves significantly better results on both DnCNN and MemNet without any hyperparameter tuning demonstrating its practical usefulness. The results are presented on the CIFAR-$10$ dataset with noise strength $\sigma$ = $0.25$ and training set size equal to $1\%$ of the entire trainset.}
\label{fig:diff_arch}
\end{figure}

We adopted the denoiser network proposed by~\cite{zhangdncnn} i.e. DnCNN for all our ablations in the main draft. In order to demonstrate the adaptability and effectiveness of our proposed approach (DE-CROP) across diffrent choices of denoiser architecture, we compare our performance against Salman \etal~\cite{salman2020denoised} on a different denoiser architecture i.e. MemNet~\cite{tai2017memnet}. Similar to the results on DnCNN in the main draft (refer Fig. $4$ in the main draft), we significantly outperform Salman \etal \cite{salman2020denoised} in a limited data setting ($D_{train}^{lim}$ = $1\%$ of $D_{train}$) across all the radii on CIFAR-$10$ with noise strength $\sigma$ = $0.25$ (refer Fig.~\ref{fig:diff_arch}).

\vspace{0.2in}
\vspace{0.2in}
\section{Effect of increasing interpolated samples}
As described in Sec. $4.1$ of the main draft, we generate one boundary sample ($x_{b}^{i}$) and one interpolated sample ($x_{int}^{i}$) for each data sample ($x_{o}^{i}$) belonging to our available limited training set ($D_{train}^{lim}$). In this section, we aim to explore whether increasing the amount of interpolated samples generated per sample leads to proportional benefit in the overall certification performance. In Fig.~\ref{fig:multi_inter} we observe that increasing the amount of interpolated samples generated (per sample) from $1$ (also reported in the main draft) to $2$ and $5$ does not increase the performance. Thus, we generate $1$ interpolated sample only as it provides similar performance with cheaper computational overhead. Note, our performance with any of the three choices discussed ($1$, $2$ or $5$ interpolated samples) comfortably outperforms other state-of-the-art techniques.
\newpage
\begin{figure}[htp]
\label{fig:multi_inter}
\centering
\centerline{\includegraphics[width=0.7\textwidth]{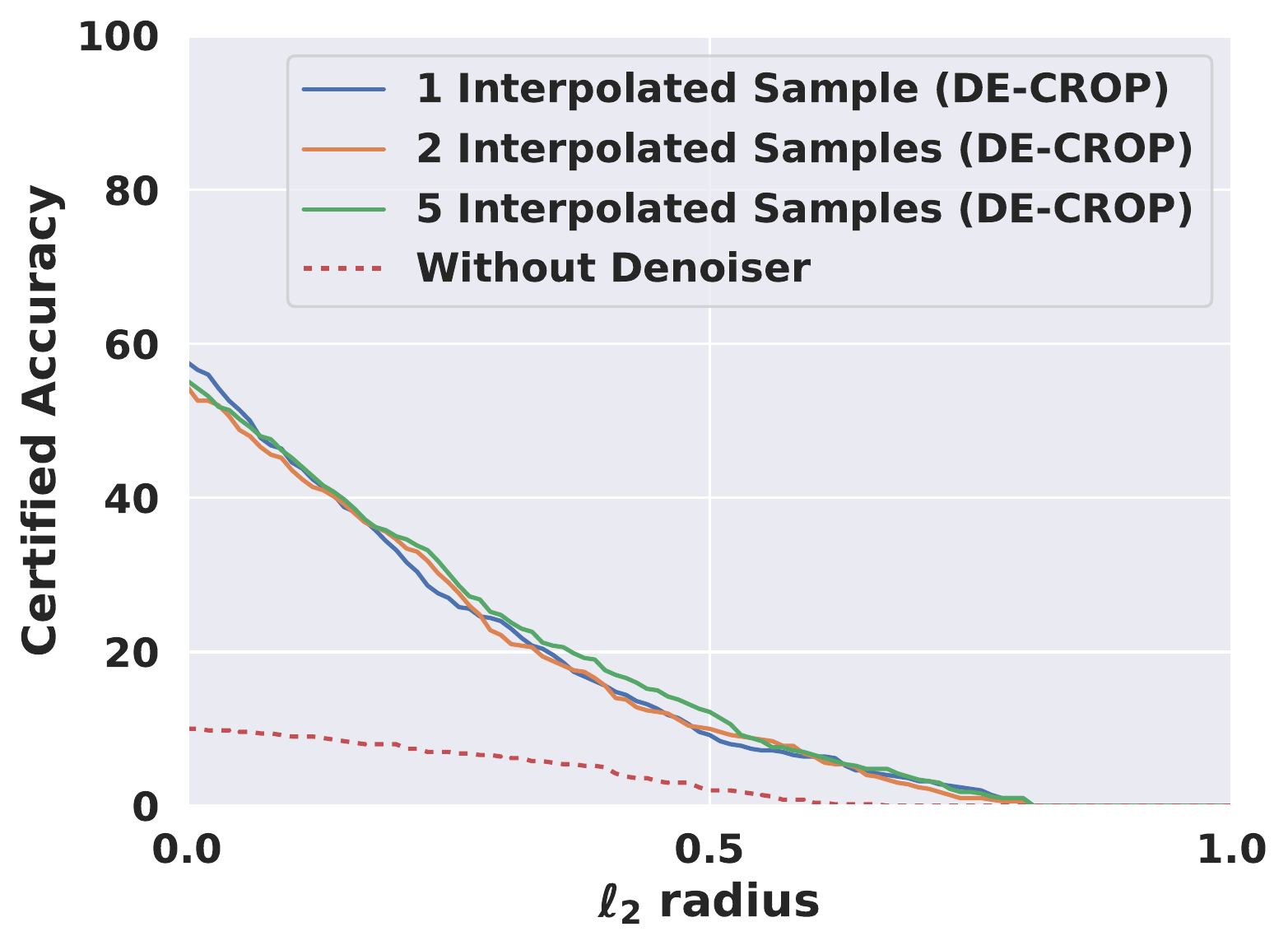}}
\caption{Effect of increasing the number of interpolated samples generated per sample on certification performance for CIFAR-$10$ dataset with noise strength $\sigma$ = $0.25$ and available limited training set = $1\%$ of entire training data. We observe that the improvement in performance is not proportional to the number of interpolated samples generated. Hence, we generate only $1$ interpolated samples in our proposed approach DE-CROP.}
\label{fig:multi_inter}
\end{figure}

\section{Performance of DE-CROP when boundary samples crafted via different adversarial attacks}

\begin{table}[htp]
\centering
\begin{tabular}{|c|c|ccc|}
\hline
\multirow{2}{*}{\textbf{Method}} & \textbf{Standard Certified} & \multicolumn{3}{c|}{\textbf{Robust Certified}} \\ \cline{2-5} 
                      & (r=0.00) & \multicolumn{1}{c|}{(r=0.25)} & \multicolumn{1}{c|}{(r=0.50)} & (r=0.75) \\ \hline
Baseline              & 29.80    & \multicolumn{1}{c|}{9.20}     & \multicolumn{1}{c|}{1.40}     & 0.00     \\ \hline
DE-CROP (PGD)         & 57.60    & \multicolumn{1}{c|}{27.20}    & \multicolumn{1}{c|}{9.20}     & 2.20     \\ \hline
DE-CROP (Auto Attack) & 53.20    & \multicolumn{1}{c|}{29.20}    & \multicolumn{1}{c|}{11.20}    & 1.20     \\ \hline
DE-CROP (DeepFool)    & 54.40    & \multicolumn{1}{c|}{26.40}    & \multicolumn{1}{c|}{10.20}    & 1.80     \\ \hline
\end{tabular}
\caption{Investigating the effect of the choice of adversarial attack (for crafting boundary samples) on certified performance using our proposed approach DE-CROP.  We significantly outperform the baseline across the different choices of adversarial attacks (i.e., PGD, DeepFool, and Auto Attack). We observe that although all three of them perform similarly, PGD obtains better certified standard accuracy compared to Auto Attack and DeepFool. Thus, we use PGD in the main draft.}
\label{tab:my-table}
\end{table}
\vspace{-0.1in}
\section{Importance of boundary samples in DE-CROP}
\begin{table}[htp]
\centering
\begin{tabular}{|c|c|ccc|}
\hline
\multirow{2}{*}{\textbf{Method}} & \textbf{Standard Certified} & \multicolumn{3}{c|}{\textbf{Robust Certified}}                \\ \cline{2-5} 
         & (r=0.00) & \multicolumn{1}{c|}{(r=0.25)} & \multicolumn{1}{c|}{(r=0.50)} & (r=0.75) \\ \hline
Baseline & 29.80    & \multicolumn{1}{c|}{9.20}     & \multicolumn{1}{c|}{1.40}     & 0.00     \\ \hline
DE-CROP without Boundary Samples & 52.46                       & \multicolumn{1}{c|}{20.26} & \multicolumn{1}{c|}{5.80} & 0.40 \\ \hline
DE-CROP with Boundary samples    & \textbf{57.60}                       & \multicolumn{1}{c|}{\textbf{27.20}} & \multicolumn{1}{c|}{\textbf{9.20}} & \textbf{2.20} \\ \hline
\end{tabular}
\caption{Demonstrating the utility of crafting boundary samples for our proposed approach: DE-CROP. In absence of boundary samples, interpolated logits are generated by using a pair of random samples from the same class (average results over three runs are reported). In contrast, we observe that the certified performance significantly improves when we utilize adversarial attack to generate boundary samples and use them along with the original samples to obtain interpolated logits. Thus, clearly highlighting the importance of crafting boundary samples in our method (DE-CROP) for certified defense in limited data.}
\label{tab:my-table}
\end{table}

\end{document}